\documentclass[conference]{IEEEtran}
\IEEEoverridecommandlockouts
\usepackage{cite}
\usepackage{amsmath,amssymb,amsfonts}
\usepackage{algorithmic}
\usepackage{graphicx}
\usepackage{textcomp}
\usepackage{xcolor}
\usepackage{multirow}
\usepackage{threeparttable}
\usepackage{makecell}
\usepackage[colorlinks=true, urlcolor=blue]{hyperref}
\def\BibTeX{{\rm B\kern-.05em{\sc i\kern-.025em b}\kern-.08em
    T\kern-.1667em\lower.7ex\hbox{E}\kern-.125emX}}
\begin{document}

\title{A Unified Biomedical Named Entity Recognition Framework with Large Language Models\\
}

\author{\IEEEauthorblockN{
Tengxiao Lv\textsuperscript{1}, 
Ling Luo\textsuperscript{1,*},
Juntao Li\textsuperscript{1}, 
Yanhua Wang\textsuperscript{2},
Yuchen Pan\textsuperscript{1},
Chao Liu\textsuperscript{1}, 
Yanan Wang\textsuperscript{1},\\
Yan Jiang\textsuperscript{3},
Huiyi Lv\textsuperscript{3},
Yuanyuan Sun\textsuperscript{1},
Jian Wang\textsuperscript{1},
Hongfei Lin\textsuperscript{1}
\\ 
\IEEEauthorblockA{
\textsuperscript{1}College of Computer Science and Technology, Dalian University of Technology, Dalian, China \\
\textsuperscript{2}Air Force Communications NCO Academy, Dalian, China \\
\textsuperscript{3}Department of Pharmacy, Second Affiliated Hospital of Dalian Medical University, Dalian, China \\
\textsuperscript{*}To whom correspondence should be addressed: lingluo@dlut.edu.cn
}
}}
\maketitle

\begin{abstract}
Accurate recognition of biomedical named entities is critical for medical information extraction and knowledge discovery. However, existing methods often struggle with nested entities, entity boundary ambiguity, and cross-lingual generalization. In this paper, we propose a unified Biomedical Named Entity Recognition (BioNER) framework based on Large Language Models (LLMs). We first reformulate BioNER as a text generation task and design a symbolic tagging strategy to jointly handle both flat and nested entities with explicit boundary annotation. To enhance multilingual and multi-task generalization, we perform bilingual joint fine-tuning across multiple Chinese and English datasets. Additionally, we introduce a contrastive learning-based entity selector that filters incorrect or spurious predictions by leveraging boundary-sensitive positive and negative samples. Experimental results on four benchmark datasets and two unseen corpora show that our method achieves state-of-the-art performance and robust zero-shot generalization across languages. The source codes are freely available at \href{https://github.com/dreamer-tx/LLMNER}{https://github.com/dreamer-tx/LLMNER}.
\end{abstract}

\begin{IEEEkeywords}
Biomedical Texts, Named Entity Recognition, Large Language Models
\end{IEEEkeywords}

\section{Introduction}
Biomedical texts, such as electronic medical records and medical textbooks, contain a large amount of critical information. These texts are essential for improving clinical decision-making and supporting biomedical research. However, their irregular and unstructured format limits their direct applicability in medical information systems. Therefore, extracting structured information—particularly Biomedical Named Entity Recognition (BioNER)—has become a key challenge~\cite{cariello2021comparison}.

\begin{figure}[t]
\centerline{\includegraphics[width=0.9\columnwidth]{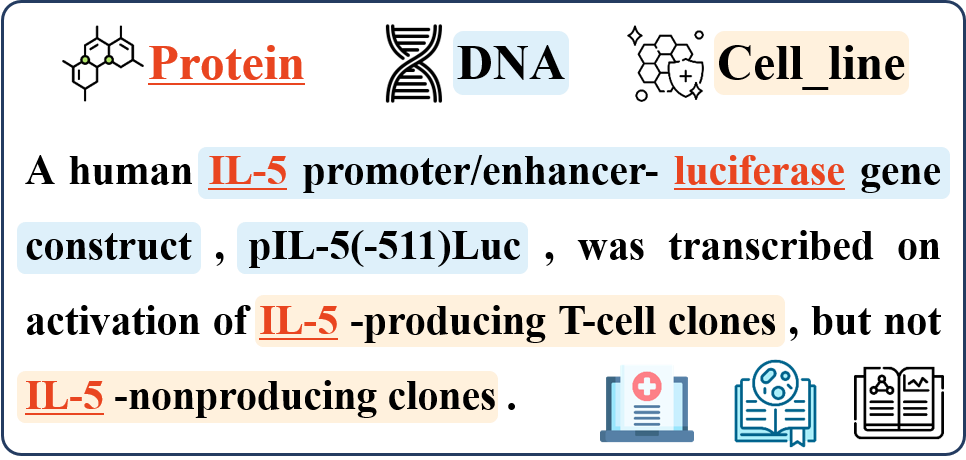}}
\caption{An example of the BioNER task from the GENIA dataset.}
\label{qiantaoshiti-yangli}
\end{figure}

The BioNER task is complex due to several factors. Biomedical texts often include highly specialized and ambiguous terms, making it difficult to determine precise entity boundaries. In addition, nested entities are frequently encountered in complex expressions. For example, as shown in Fig.~\ref{qiantaoshiti-yangli}, the phrase “IL-5 promoter/enhancer-luciferase gene construct” is annotated as a DNA entity that contains two nested Protein entities: “IL-5” and “luciferase”. Both flat and nested entities often appear in the same sentence, increasing the difficulty of modeling. Furthermore, differences across languages introduce further challenges. Chinese and English differ significantly in terminology construction, sentence structure, and expression style, requiring language-specific modeling approaches~\cite{tsuruoka2003boosting}.

To facilitate the development of BioNER, significant efforts have been made. Early statistical models such as Hidden Markov Models (HMM)~\cite{zhao2004named}, Support Vector Machines (SVM)~\cite{makino2002tuning}, and Conditional Random Fields (CRF)~\cite{settles2005abner}, often combined with hand-crafted features, achieved limited success~\cite{luo2020chinese}, but struggled with nested and multilingual settings. Deep learning models like BiLSTM-CRF~\cite{luo2018attention} and Lattice-LSTM~\cite{zhang2018chinese} improved contextual representation but lacked a unified framework for handling diverse entity structures across languages.

The advent of pre-trained language models (PLMs) marked a major milestone in BioNER. Models such as BioBERT~\cite{lee2020biobert} brought significant improvements, and hybrid BERT-based architectures with CRF or span-based decoders have become widely adopted~\cite{chen2022named}. To better handle nested entities, recent methods like W2NER~\cite{li2022unified}, CNNNER~\cite{yan2023embarrassingly}, and DiFiNet~\cite{cai2024difinet} have introduced span-based feature matrices, relation modeling, and boundary-aware learning strategies. Nevertheless, these methods heavily rely on complex structural designs or annotation strategies~\cite{zhao2023chinese} and are typically designed for a specific language or dataset, limiting their ability to generalize. AIONER~\cite{luo2023aioner} has attempted to develop a unified model trained on multiple BioNER datasets, showing promise but still faces challenges in zero-shot scenarios and with nested entity recognition. Overall, these PLM methods fall short of offering a comprehensive solution for unified BioNER that can jointly support flat and nested entities, across English and Chinese within a framework.

\begin{figure*}[t]
\centerline{\includegraphics[width=1.0\textwidth]{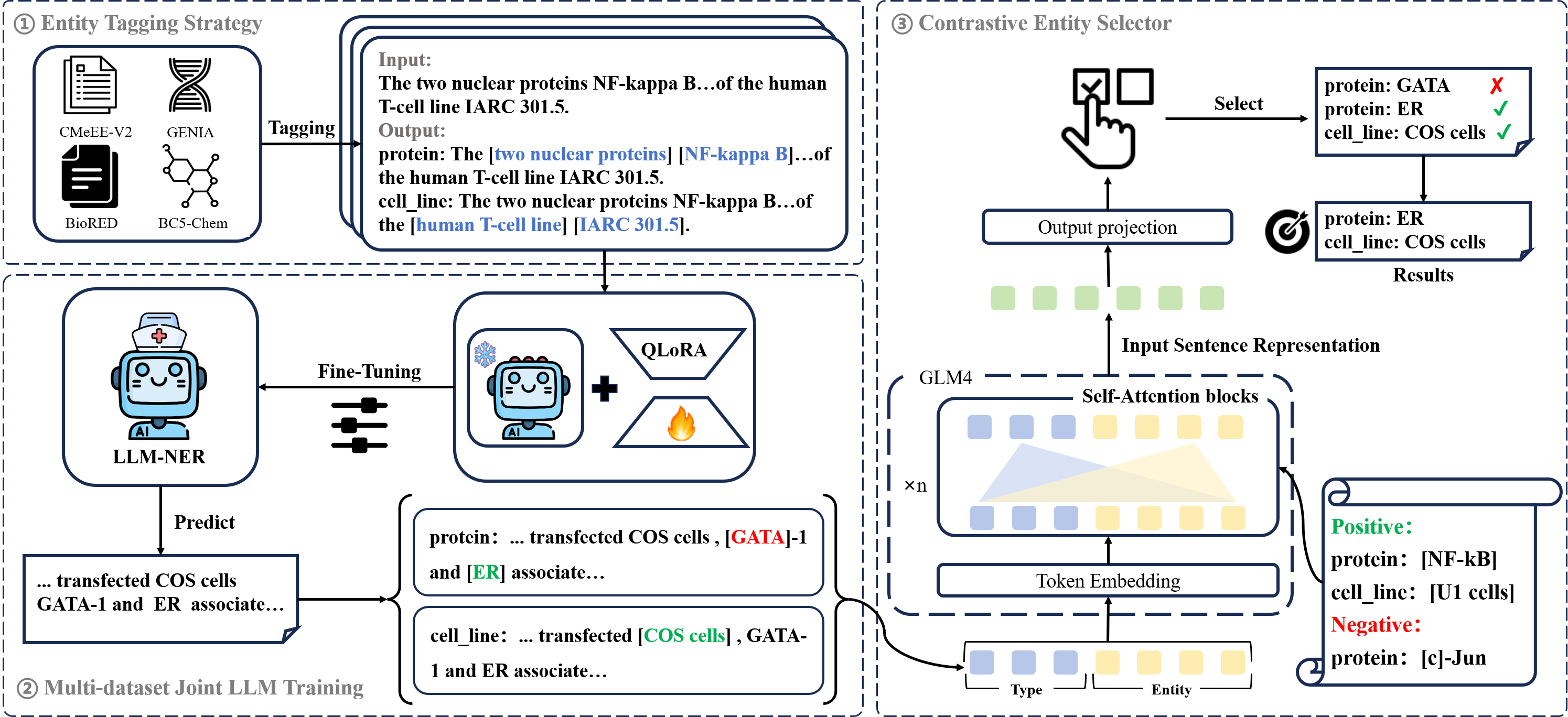}}
\caption{Overview of our proposed BioNER framework. Blue represents entities in the training set, while green indicates correctly predicted entities and red indicates incorrectly predicted entities.}
\label{zhengtikuangjia}
\end{figure*}

Large Language Models (LLMs) have recently shown great potential in BioNER due to their deep semantic understanding capabilities~\cite{luo2024taiyi}. Instruction-tuned frameworks such as UniversalNER~\cite{zhou2023universalner} leverage LLMs to build general-purpose NER systems via dataset distillation. However, most LLM-based methods only generate entity types and names, without providing their exact entity mention positions in the text~\cite{wang2023instructuie}, which limits their practical utility in biomedical information extraction, where mention-level accuracy is critical.

To address these challenges, we propose an enhanced LLM-based BioNER method that reformulates the task as a text generation task using a symbolic tagging strategy. This enables unified recognition across both Chinese and English, as well as for flat and nested entities. We first design and compare various entity tagging strategies, and find that the symbolic tagging scheme with effective boundary representation yields superior performance. Then, we employ a multi-dataset joint fine-tuning approach to enhance the model’s generalization across tasks and languages. Finally, a contrastive fine-tuning selector is introduced to filter out incorrect entities and further improve overall performance.

Our main contributions are as follows:

\begin{itemize}
  \item We propose a unified generative framework for BioNER that supports the recognition of both flat and nested entities in Chinese and English. Several entity tagging strategies are designed and compared, specifically tailored to LLMs. The framework is jointly fine-tuned on multiple BioNER datasets and extensively evaluated on various open-source LLMs, including LLaMA~\cite{dubey2024llama}, GLM~\cite{glm2024chatglm}, Qwen~\cite{yang2024qwen2}, and DeepSeek~\cite{guo2025deepseek}.
  
  \item We introduce a contrastive entity selector that leverages fine-tuning with positive and negative sample to effectively filter out candidate entities with incorrect boundaries or types, thereby enhancing recognition precision.

  \item Extensive experiments are conducted on four Chinese and English BioNER datasets. The results demonstrate that our proposed method can accurately identify both flat and nested entities along with their exact spans, achieving competitive performance. Furthermore, the model exhibits strong zero-shot generalization capability when evaluated on unseen datasets.
\end{itemize}

\section{METHODS}

The overview of our proposed framework is as illustrated in Fig.~\ref{zhengtikuangjia}. The framework reformulates BioNER as a text generation task, enabling unified recognition of flat and nested entities across both Chinese and English texts. To support this, we first explore different entity tagging strategies and adopt a symbolic tagging approach that effectively represents entity boundaries in generative outputs. Building on this, we perform joint fine-tuning on multiple Chinese and English BioNER datasets to improve the model’s generalization across tasks and languages. Finally, to further enhance precision and reduce boundary-related errors, we introduce a contrastive entity selector that filters out incorrect or spurious predictions. Together, these components form a unified and robust BioNER framework tailored for LLMs.

\subsection{Entity Tagging Strategy}\label{sec:tagging_strategy}

Disambiguating the meaning of entities in biomedical texts often relies on contextual information. Compared to general-domain NER, biomedical applications typically demand higher precision and reliability. Providing the exact start and end positions of each recognized entity helps users trace the original context more effectively, which increases the practical utility of BioNER systems but also makes the task more complex.

\begin{figure}[t]
\centerline{\includegraphics[width=\columnwidth]{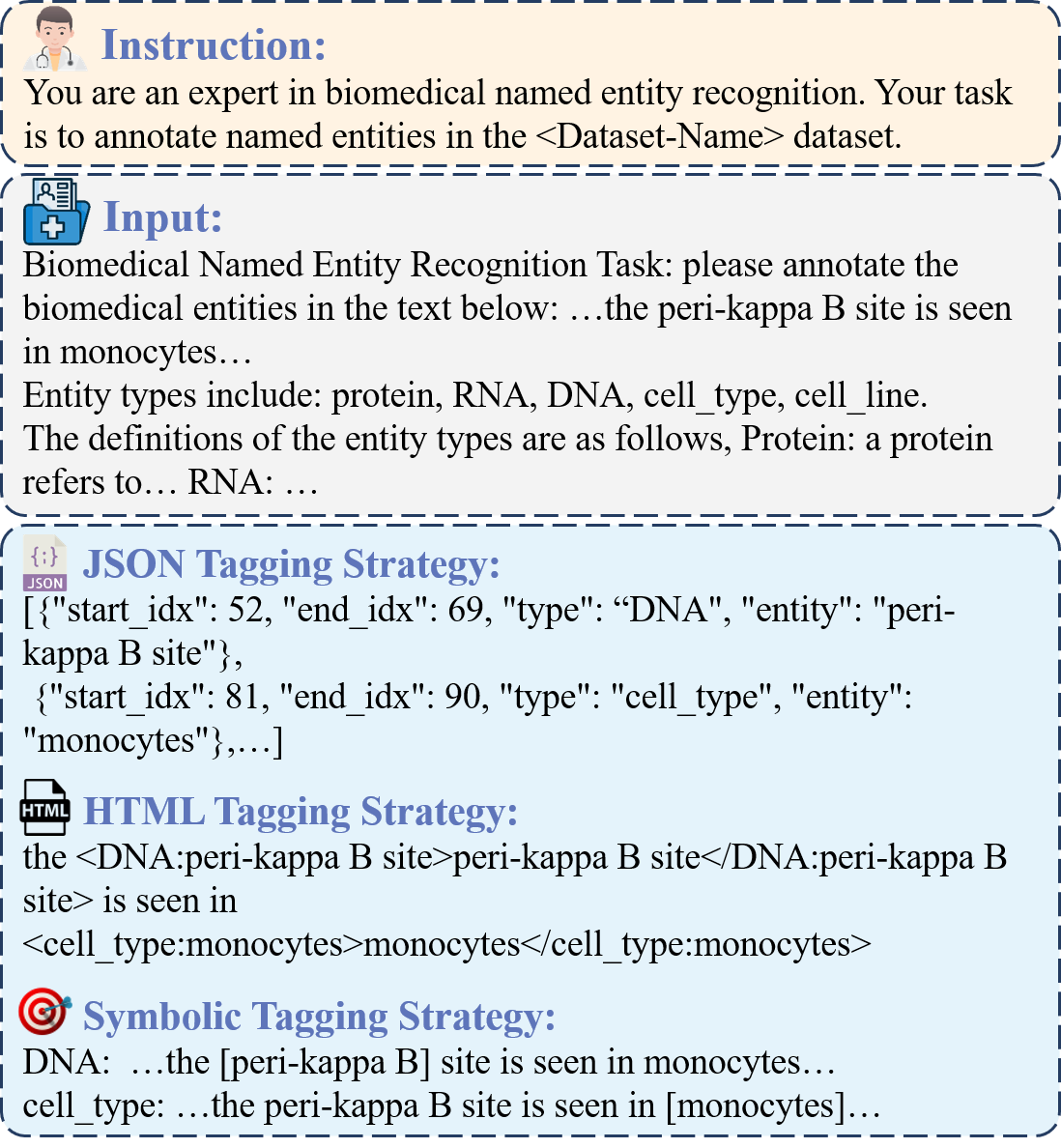}}
\caption{Illustration of different entity tagging strategies. 
         Example adapted from the GENIA dataset; 
         the {\rmfamily\textless Dataset-Name\textgreater} placeholder in instructions 
         varies according to specific datasets.}
\label{biaozhucelue-prompt}
\end{figure}

To address the needs of the BioNER task, we propose three entity tagging strategies inspired by existing LLM-based NER output methods: a JSON tagging strategy, an HTML tagging strategy, and a symbolic tagging strategy. Example prompts are illustrated in Fig.~\ref{biaozhucelue-prompt}. The input first designates the model as an expert in BioNER, then provides the original sentence along with definitions for each entity type. The output is generated in accordance with the selected tagging strategy.

\textbf{JSON Tagging Strategy}: Entity information is represented in a structured key-value format, containing the start and end index positions in the original sentence (start\_idx, end\_idx), the entity type (type), and the entity text (entity).

\textbf{HTML Tagging Strategy}: The model annotates named entities within the sentence using custom HTML-like tags that include both the entity type and the exact entity text. Specifically, {\rmfamily\textless type:entity\textgreater} marks the start, and {\rmfamily\textless/type:entity\textgreater} marks the end of an entity. In contrast, BioNER-Llama~\cite{keloth2024advancing} uses {\rmfamily\textless mask\textgreater\textless/mask\textgreater} tags to annotate entities. Our method embeds both type and entity text in the tag itself, which facilitates accurate matching, especially in scenarios involving nested and multi-type entities.

\textbf{Symbolic Tagging Strategy}: In this method, the output first specifies the entity type, followed by the original text sequence. Special symbols “\texttt{[}” and “\texttt{]}” are used to indicate entity boundaries, where “\texttt{[}” marks the start and “\texttt{]}” marks the end. Entities are tagged separately by type, which avoids nesting within the same type and thus simplifies both annotation and model learning.

By designing these tagging strategies, we reformulate the traditional sequence labeling task of BioNER into a text generation task. This transformation aligns well with the autoregressive nature of LLMs and lays the foundation for their effective application in biomedical entity recognition.


\subsection{Multi-dataset Joint LLM Training}

To further enhance the generalization capability of the model in multilingual and multi-dataset BioNER tasks, we propose a multi-dataset fine-tuning approach built upon our entity tagging strategy. This approach involves joint training on multiple Chinese and English BioNER datasets, leveraging task diversity to help the model learn more generalized feature representations and improve its adaptability in zero-shot and cross-dataset scenarios.

Specifically, we select four BioNER datasets: the Chinese dataset CMeEE-V2~\cite{zhang2021cblue}, and the English datasets GENIA~\cite{kim2003genia}, BioRED~\cite{luo2022biored}, and BC5CDR-Chemical~\cite{li2016biocreative}. CMeEE-V2 and GENIA include nested entities, whereas BioRED and BC5CDR-Chemical contain flat entities, revealing clear structural differences. To fully utilize the complementary information of these heterogeneous datasets, we adopt a unified symbolic output scheme for joint modeling and instruction fine-tuning the LLMs within a multi-dataset framework.

For prompt design, each training instance explicitly specifies the dataset name and dynamically adjusts the entity type descriptions included in the prompt, as shown in Fig.~\ref{biaozhucelue-prompt}. For example, when processing the CMeEE-V2 dataset, the model is provided only with the nine entity type definitions specific to that dataset. In the case of GENIA, those definitions are replaced with its five corresponding categories. This differentiated prompting strategy allows the model to accurately comprehend the semantic context and output expectations of each task, thereby reducing cross-task interference.

Additionally, to prevent overfitting to any single task, the four datasets are mixed at the sentence level and dynamically shuffled during training to ensure balanced learning across tasks. Furthermore, to accommodate multilingual settings, Chinese and English samples are alternated in the training corpus, allowing the model to develop balanced semantic representation capabilities in both languages.

To efficiently adapt LLMs to these diverse training objectives, we employ the QLoRA method~\cite{dettmers2023qlora} for parameter-efficient fine-tuning. QLoRA enables effective training of large models using limited hardware resources while preserving their original language understanding and generation capabilities.

\subsection{Contrastive Entity Selector}

An error analysis of fine-tuned LLMs on the BioNER task serves as the foundation for designing an entity selector aimed at addressing observed shortcomings. Specifically, most recognition errors involve mismatches in entity boundaries or type. These errors can be categorized into four types based on different matching conditions: (1) correct position but incorrect type (\textit{Type}); (2) partially overlapping span with correct type (\textit{Span}); (3) partially overlapping span with incorrect type (\textit{Type\&Span}); and (4) completely mismatched span (\textit{Spurious}).

Experimental results indicate that boundary-related errors comprise a substantial portion of overall recognition errors, with most boundary deviations occurring within two tokens. This reveals a significant limitation of LLMs in accurately identifying entity boundaries. However, biomedical applications demand high precision in NER, particularly in high-stakes scenarios such as clinical practice.

To address this issue, we propose a contrastive sample-based entity selector that enhances the model’s sensitivity to entity boundaries and overall recognition accuracy, without relying on external knowledge. This selector builds upon the symbolic tagging strategy ultimately adopted in this work. The training data consists of positive and negative samples constructed as follows:

Labeled entities from the training set are used as positive samples. Negative samples are generated by randomly shifting the start or end positions by ±1 or ±2 tokens or by altering the entity type. A total of 10,000 positive and negative samples are randomly selected to form the fine-tuning dataset.

Based on this dataset, we fine-tune the GLM4 model and add a classification layer to its self-attention module to determine whether a candidate entity is valid. The detailed process is as follows:

Let the original text sequence be $x = [x_1, x_2, ..., x_n]$, where $i$ and $j$ denote the start and end positions of a candidate entity in the text ($0 \leq i \leq j \leq n$). Special tokens are inserted before and after the candidate entity to generate a new input sequence $x'$:
\begin{equation}
x' = [x_{0:(i-1)} \oplus s_1 \oplus x_{i:j} \oplus s_2 \oplus x_{(j+1):n}]
\end{equation}
where $s_1$ and $s_2$ are special tokens marking the start and end of the entity, respectively, and $\oplus$ represents sequence concatenation.

The sequence $x'$ is passed into the GLM4 model for contextual encoding:
\begin{equation}
h = \text{SelfAttention}(t \oplus x')
\end{equation}
where $\text{SelfAttention}$ denotes the multi-layer self-attention mechanism, $t$ is the entity type, and $h$ is the output hidden representation of the sequence.

Finally, the hidden state $h$ is used to predict the validity of the candidate entity:
\begin{equation}
o = \text{Sigmoid}\left(\text{FC}_{\text{cls}}(h)\right)
\end{equation}
where $\text{FC}_{\text{cls}}$ is a fully connected classification layer, and $o$ is the probability score indicating whether the candidate entity is valid.

Using this approach, we implement an entity selector which, when applied to the candidate entities produced by LLMs, filters out those with boundary or type errors. This enhances the reliability of BioNER and better meets the precision requirements of biomedical applications.

\section{EXPERIMENTS}

\subsection{Experimental Datasets and Settings}

This study evaluates the proposed method on two Chinese BioNER datasets and four English datasets. The Chinese datasets include CMeEE-V2~\cite{zhang2021cblue} and CCKS2019-Anatomical Site (CCKS2019-AS)~\cite{han2020overview}, while the English datasets consist of GENIA~\cite{kim2003genia}, BioRED~\cite{luo2022biored}, BC5CDR-Chemical (BC5-Chem)~\cite{li2016biocreative}, and NCBI-Disease~\cite{dougan2014ncbi}. The statistics of these datasets are summarized in Table~\ref{Statistics}. Among them, CMeEE-V2 and GENIA are annotated with nested entities, whereas the remaining four datasets contain only flat entities. To evaluate the model’s generalization capabilities, CCKS2019-AS and NCBI-Disease are treated as unseen test sets in the multi-dataset learning setting.

Since CMeEE-V2 does not provide a standard test set, we follow previous practice and randomly select 500 samples from its development set to serve as the evaluation set. All datasets are preprocessed into a unified input-output format to align with our generation-based tagging strategy.

Evaluation Metrics: We employ Precision (P), Recall (R), and F1 score (F1) as standard evaluation metrics. A predicted entity is considered correct only if both its span and type exactly match the gold annotation.

Hyperparameter Settings: All experiments are conducted on a single NVIDIA L40 GPU. Models are trained for 15 epochs with a batch size of 6 and a learning rate of 2e-4. A dropout rate of 0.05 is applied, and the AdamW optimizer is used.


\begin{table}[t]
\centering
\caption{The BioNER datasets used in our study}
\resizebox{\linewidth}{!}{
\begin{tabular}{lcccccc}
\hline
Dataset         & Language & Nesting & \makecell[c]{Type\\Count} & \makecell[c]{Entity\\Count} & \makecell[c]{Train\\Size} & \makecell[c]{Test\\Size} \\ \hline
CMeEE-V2        & Ch       & Yes     & 9          & 82,649 & 15,000 & 500 \\
CCKS2019-AS     & Ch       & No      & 1          & 17,653 & 1,000  & 379 \\
GENIA           & En       & Yes     & 5          & 50,509 & 16,692 & 1,854 \\
BioRED          & En       & No      & 6          & 1,688  & 5,507  & 1,106 \\
BC5-Chem     & En       & No      & 1          & 10,550 & 9,141  & 4,797 \\
NCBI-Disease    & En       & No      & 1          & 5,921  & 6,347  & 940 \\ \hline
\end{tabular}
}
\label{Statistics}
\end{table}

\begin{table}[t]
\centering
\caption{Performance Comparison of Three Entity Tagging Strategies}
\resizebox{\linewidth}{!}{
\begin{tabular}{lccccccc}
\hline
         & \multicolumn{3}{c}{CMeEE-V2} &  & \multicolumn{3}{c}{GENIA} \\ \cline{2-4} \cline{6-8} 
         & P        & R       & F1      &  & P       & R      & F1     \\ \hline
JSON     & 70.86    & 77.26   & 73.92   &  & 81.21   & 72.47  & 76.59  \\
HTML     & 69.21    & 71.43   & 70.30   &  & 79.22   & 72.92  & 75.94  \\
Symbolic & \textbf{73.66}    & \textbf{78.81}   & \textbf{76.15}   &  & \textbf{81.96}   & \textbf{77.41}  & \textbf{79.62}  \\ \hline
\end{tabular}
}
\label{Three-Entity-Tagging-Strategies}
\end{table}

\subsection{Performance of Different Entity Tagging Strategies}

As described in Section~\ref{sec:tagging_strategy}, three entity tagging strategies were designed for LLMs. To identify the most effective strategy, we fine-tuned the GLM4-9B-Chat model using each tagging strategy on the Chinese CMeEE-V2 task and the English GENIA task for comparison. The results are shown in Table~\ref{Three-Entity-Tagging-Strategies}.

The results show that the symbolic tagging strategy achieves the highest F1 score on both datasets, clearly outperforming the JSON and HTML formats. Specifically, the HTML tagging strategy yields an F1 score of only 70.30\% on CMeEE-V2. This may be due to the complex sentence structures and frequent nested entities in the dataset, which make the HTML tagging format structurally complicated. As a result, generative models find it difficult to learn effectively, often leading to parsing errors or missing entities during generation. The JSON strategy also underperforms, likely because LLMs struggle to accurately generate and interpret the numerical indices used to indicate entity boundaries. In contrast, the symbolic tagging strategy simplifies the annotation format by separating entity types and using lightweight boundary symbols, which reduces annotation ambiguity and structural complexity. This design aligns well with the generation process of LLMs and is better suited for recognizing both flat and nested entities while providing precise span information.

\subsection{Performance Comparison of Different LLMs}

\begin{figure}[t]
\centerline{\includegraphics[width=1.0\columnwidth]{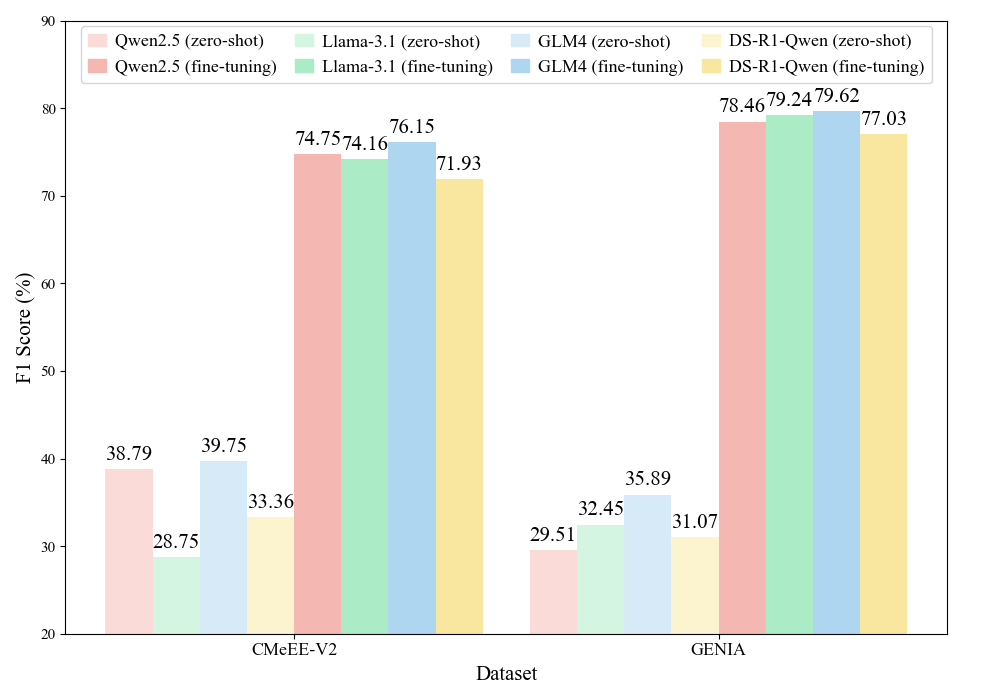}}
\caption{Zero-shot and fine-tuning performance of different LLMs.}
\label{butongjizuomoxing}
\end{figure}

Based on the results of previous experiments, the best-performing symbolic tagging strategy is selected for further evaluation. This section compares the zero-shot and fine-tuning performance of four different LLMs on the Chinese dataset CMeEE-V2 and the English dataset GENIA. The evaluated models include Qwen2.5-7B-Instruct, Llama-3.1-8B-Instruct, GLM4-9B-Chat, and DeepSeek-R1-Distill-Qwen-7B (DS-R1-Qwen). The corresponding F1 scores are shown in Fig.~\ref{butongjizuomoxing}.

From Fig.~\ref{butongjizuomoxing}, we observe that all four LLMs exhibit relatively low F1 score under zero-shot conditions, suggesting that general-purpose LLMs perform poorly on domain-specific tasks such as BioNER without additional adaptation. After fine-tuning, all models demonstrate notable improvements, with GLM4-9B-Chat achieving the highest F1 score. Interestingly, the reasoning-oriented DeepSeek distilled model underperforms after fine-tuning, even falling behind the original Qwen model. This may be due to its training objectives being more focused on logical reasoning, which might not align well with the span-level precision and boundary sensitivity required by the BioNER task. Overall, these results indicate that fine-tuning is essential for adapting general LLMs to BioNER and further demonstrate the effectiveness of combining the symbolic tagging strategy with fine-tuning.

\begin{figure}[t]
\centerline{\includegraphics[width=0.9\columnwidth]{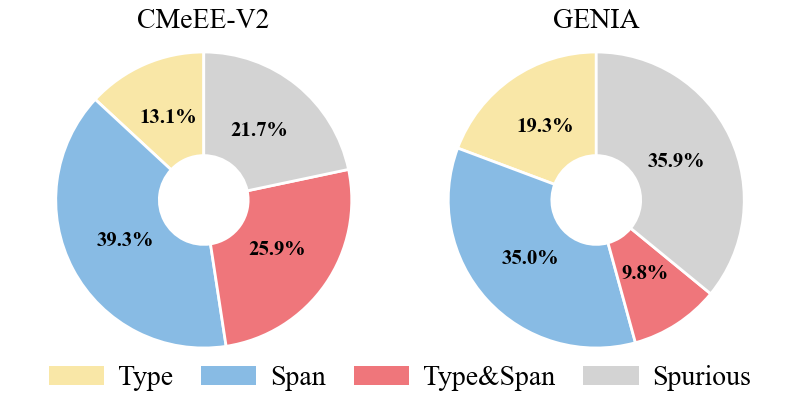}}
\caption{Summary of error analysis. These errors can be categorized into four types based on different matching conditions: (1) correct position but incorrect type (Type); (2) partially overlapping span with correct type (Span); (3) partially overlapping span with incorrect type (Type\&Span); (4) completely mismatched span (Spurious).}
\label{cuowuleixingzhanbi}
\end{figure}

Further analysis is conducted on the error types and boundary-related errors in the outputs of fine-tuned models. Fig.~\ref{cuowuleixingzhanbi} illustrates the distribution of recognition error types for the best-performing GLM4-9B-Chat model using the symbolic tagging strategy on the CMeEE-V2 and GENIA datasets. 

As shown in Fig.~\ref{cuowuleixingzhanbi}, errors related to entity boundary overlap, specifically the "Span" and "Type\&Span" categories, constitute a significant portion of recognition errors, accounting for 65.2\% in the CMeEE-V2 dataset. This reveals that the current model still exhibits substantial limitations in accurately identifying entity boundaries and types.

\begin{figure}[t]
\centerline{\includegraphics[width=0.9\columnwidth]{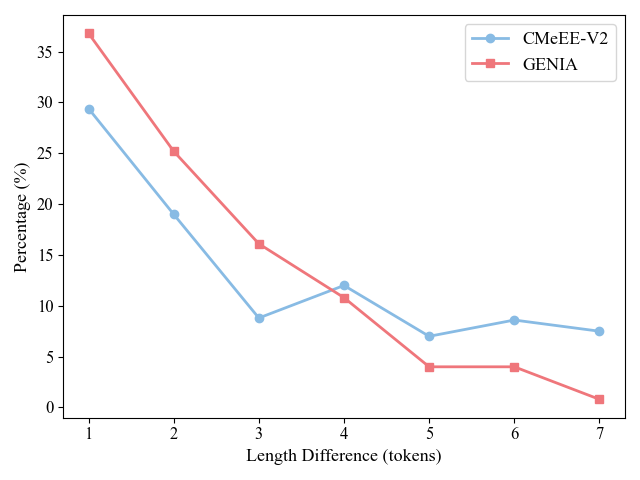}}
\caption{Token-level deviation distribution of boundary errors.}
\label{token-changduchazhi}
\end{figure}

To further investigate the nature of boundary-related errors, Fig.~\ref{token-changduchazhi} provides a detailed token-level analysis of length differences associated with these mistakes.

As illustrated in Fig.~\ref{token-changduchazhi}, approximately 50\% of boundary errors occur within a deviation range of two tokens, indicating that the model often makes slight misjudgments near entity boundaries. This observation further underscores the necessity of enhancing the model's boundary sensitivity. To address this issue, we propose a contrastive entity selector specifically designed to reduce boundary ambiguity and filter out invalid entity predictions.

\subsection{Effect of the Contrastive Entity Selector}
To verify the effectiveness of the selector based on contrastive learning with positive and negative samples, we compare three different selection mechanisms on the BioNER task: a fine-tuned BERT classifier (BERT-CLS), an autoregressive selector built upon the GLM4-9B-Chat (LLM-CAUSAL), and the proposed classification-layer selector based on GLM4-9B-Chat (LLM-CLS). Their performance, along with that of the base model, on the CMeEE-V2 and GENIA datasets is summarized in Table~\ref{Three-Different-Entity-Selectors}. The baseline model is GLM4-9B-Chat fine-tuned with the symbolic tagging strategy.

\begin{table}[t]
\centering
\caption{Performance Comparison of Three Different Entity Selectors}
\resizebox{\linewidth}{!}{
\begin{tabular}{lccccccc}
\hline
            & \multicolumn{3}{c}{CMeEE-V2} &  & \multicolumn{3}{c}{GENIA} \\ \cline{2-4} \cline{6-8} 
            & P        & R       & F1      &  & P       & R      & F1     \\ \hline
Baseline    & 73.66    & \textbf{78.81}   & 76.15   &  & 81.96   & \textbf{77.41}  & 79.62  \\
+BERT-CLS   & 74.27    & 78.68   & 76.41   &  & 82.40   & 77.06  & 79.64  \\
+LLM-CAUSAL & 74.92    & 78.76   & 76.79   &  & 84.05   & 77.22  & 80.49  \\
+LLM-CLS    & \textbf{75.86}    & 78.73   & \textbf{77.26}   &  & \textbf{84.56}   & 77.28  & \textbf{80.76}  \\ \hline
\end{tabular}
}
\label{Three-Different-Entity-Selectors}
\end{table}

From Table~\ref{Three-Different-Entity-Selectors}, we observe that all three selectors improve the precision of the baseline model across both datasets, validating the effectiveness of contrastive selectors in the BioNER task. Among them, LLM-CLS achieves the highest F1 score on both datasets, with an improvement of approximately 1.1\% over the baseline. This demonstrates its ability to reduce recognition errors by accurately modeling consistency in entity boundaries and types, thereby improving overall recognition quality. Compared to LLM-CLS, LLM-CAUSAL achieves higher recall but lower precision on CMeEE-V2, resulting in a slightly lower F1 score. This suggests that LLM-CLS provides a better balance between precision and recall, especially in handling complex nested entities. In contrast, BERT-CLS shows only marginal improvements on both datasets, and its overall gains over the baseline are limited, indicating that although the BERT-based classifier has some discriminative power, it remains less effective than LLMs in this task setting.


\begin{table}[t]
\centering
\caption{Error Reduction via Selector}
\resizebox{\linewidth}{!}{
\begin{tabular}{lcccc}
\hline
Method   & Type  & Span  & Type\& Span & Spurious \\ \hline
Baseline & 84    & 252   & 166         & 139      \\
Ours     & 74    & 218   & 157         & 121      \\
$\Delta$        & ↓12\% & ↓13\% & 5\%        & ↓13\%    \\ \hline
\end{tabular}
}
    \begin{tablenotes}
    	\item \textbf{Notes}: Baseline denotes the error statistics of GLM4-9B-Chat fine-tuned with the symbolic tagging strategy on CMeEE-V2.
    \end{tablenotes}
\label{Reduction}
\end{table}

Table~\ref{Reduction} reports the changes in the four main error categories after applying LLM-CLS. It is evident that our method reduces the number of errors across all four types, confirming that the proposed selector effectively filters out various kinds of recognition mistakes. Notably, the most substantial improvements are observed in the “Type”, “Span”, and “Spurious” categories, each of which is reduced by over 10\%, demonstrating the method’s effectiveness in addressing boundary-related recognition issues.

\subsection{Performance of Multi-dataset Joint Training}
This experiment aims to evaluate the overall effectiveness of the proposed multi-dataset learning fine-tuning strategy in bilingual BioNER tasks. We conduct joint training and testing on the Chinese dataset CMeEE-V2 and the English datasets GENIA, BioRED, and BC5CDR-Chemical (BC5-Chem). Table~\ref{Multi-dataset} presents a performance comparison under three different settings: (1) fine-tuning on each dataset independently (Single Fine-tuning); (2) fine-tuning on all four datasets simultaneously (Joint Fine-tuning); and (3) applying the proposed selector (+LLM-CLS) on top of the jointly fine-tuned model.

\begin{table}[t]
\centering
\caption{Multi-dataset Joint Training Performance}
\resizebox{\linewidth}{!}{
\begin{tabular}{lccc}
\hline
Dataset     & Single Fine-tuning & Joint Fine-tuning & +LLM-CLS \\ \hline
CMeEE-V2    & 76.15              & 76.41             & 77.36    \\
GENIA       & 79.62              & 80.27             & 81.88    \\
BioRED      & 90.08              & 90.25             & 91.30    \\
BC5-Chem & 93.81              & 93.92             & 94.27    \\ \hline
\end{tabular}
}
\label{Multi-dataset}
\end{table}

The results in Table~\ref{Multi-dataset} demonstrate that, compared to single fine-tuning, joint fine-tuning consistently improves performance across all datasets. The observed gains in F1 score suggest that multi-dataset training enables the model to learn more generalized entity representations. Furthermore, incorporating the LLM-CLS selector on top of the joint model yields additional improvements in F1 score on all four datasets, confirming that the selector plays an effective role in enhancing entity recognition performance.

\subsection{Performance Comparison with Other Existing Methods}

To demonstrate the effectiveness of the proposed method, we compare our approach (Ours) and its variant without multi-dataset joint training (Ours w/o MDT) against a range of state-of-the-art named entity recognition models across four bilingual benchmark datasets: CMeEE-V2, GENIA, BioRED, and BC5CDR-Chemical (BC5-Chem), as shown in Table~\ref{Performance}. Due to the lack of open-source implementations for some models, certain results are reported on only a subset of datasets. The compared methods are broadly divided into two categories:

\begin{table}[]
\centering
\caption{Performance Comparison with Other Methods}
\resizebox{\linewidth}{!}{
\begin{tabular}{p{2.4cm}cccc}
\hline
                 & \multicolumn{1}{l}{CMeEE-V2} & \multicolumn{1}{l}{GENIA} & \multicolumn{1}{l}{BioRED} & \multicolumn{1}{l}{\begin{tabular}[c]{@{}l@{}}BC5-Chem\end{tabular}} \\ \hline
\multicolumn{5}{c}{BERT-based approaches}                                                                                              \\ \hline
BioBERT-CRF      & 71.93                        & 78.06                     & 88.38                      & 91.99                               \\
W2NER            & 74.23                        & 79.00                     & 88.63                      & 93.60                                \\
CNNNER           & 74.73                        & 79.13                     & 89.21                      & 92.92                               \\
DiFiNet          & 74.85                        & 79.23                     & 89.50                      & 92.45                               \\
AIONER*          & -                            & -                         & \underline{91.26}                & 92.84                               \\ \hline
\multicolumn{5}{c}{LLM-based approaches}                                                                                                       \\ \hline
ChatGPT3.5      & 47.00                        & 41.60                      & 38.10                       & 60.30                                \\
GPT-4            & 49.73                        & 48.65                     & 58.43                      & 76.39                               \\
DeepSeek-R1      & 52.18                        & 47.93                     & 58.86                      & 72.08                               \\
Taiyi*           & 65.70                         & -                         & -                          & 80.20                               \\
UniversalNER*    & -                            & 77.54                     & 88.90                       & -                                   \\ \hline
Ours w/o MDT     & \underline{77.26}                  & \underline{80.76}               & 90.96                      & \underline{93.98}                         \\
Ours             & \textbf{77.36}               & \textbf{81.88}            & \textbf{91.30}              & \textbf{94.27}                      \\ \hline
\end{tabular}
}
    \begin{tablenotes}
    	\item \textbf{Notes}: Results marked with an asterisk (*) are taken from the original paper; bold font indicates the best performance, and underlined font indicates the second-best performance. MDT refers to multi-dataset Joint Training.
    \end{tablenotes}
\label{Performance}
\end{table}


\textbf{BERT-based approaches}:
\begin{itemize}
    \item \textbf{BioBERT-CRF} utilizes BioBERT for character-level representations followed by a CRF layer for sequence labeling.
    \item \textbf{W2NER}~\cite{li2022unified} treats NER as a word-pair classification task to handle nested entities.
    \item \textbf{CNNNER}~\cite{yan2023embarrassingly} combines BERT with CNNs to model span relations for nested NER.
    \item \textbf{DiFiNet}~\cite{cai2024difinet} introduces boundary-aware contrastive modules on BioBERT to enhance nested entity recognition.
    \item \textbf{AIONER}~\cite{luo2023aioner} unifies various BioNER datasets into a common format, enabling a single model to recognize diverse entity types.
\end{itemize}

\textbf{LLM-based approaches}:
\begin{itemize}
    \item \textbf{ChatGPT-3.5, GPT-4, DeepSeek-R1} perform zero-shot NER via prompt-based inference.
    \item \textbf{Taiyi}~\cite{luo2024taiyi} adapts LLMs for BioNER through a two-stage fine-tuning process.
    \item \textbf{UniversalNER}~\cite{zhou2023universalner} constructs instruction-tuned datasets with ChatGPT and distills LLaMA-based NER models.
\end{itemize}

Experimental results in Table~\ref{Performance} demonstrate that our proposed LLM-based BioNER method achieves the highest F1 score across all four Chinese and English datasets. Our approach uses a unified model to perform multilingual and multi-dataset BioNER tasks, outperforming comparable methods such as AIONER, Taiyi, and UniversalNER, and showing superior task adaptability and stable performance.

Specifically, on nested entity datasets such as CMeEE-V2 and GENIA, our method surpasses mainstream models like CNNNER and DiFiNet, demonstrating strong capabilities in recognizing complex entity structures. On flat entity datasets such as BioRED and BC5CDR-Chemical, it also achieves the highest overall F1 score while maintaining high precision and recall, further validating the model's adaptability to entity recognition tasks with different structural characteristics.

In addition, compared with general-purpose LLMs such as GPT-4 and DeepSeek-R1, our method effectively addresses key challenges such as limited domain knowledge and weak boundary recognition by incorporating task-specific tagging strategies, multi-dataset joint training, and an entity selector. These enhancements enable instruction-tuned LLMs to achieve state-of-the-art performance on domain-specific BioNER tasks.

\subsection{Zero-Shot Generalization on Unseen Datasets}
To further validate the generalization capability of the multi-dataset learning strategy across different datasets and languages, we conduct zero-shot evaluations on two unseen BioNER datasets following bilingual multi-dataset fine-tuning: the Chinese CCKS2019-AS and the English NCBI-Disease datasets. These datasets were not included during training. The multi-dataset fine-tuned model is directly applied to these datasets without additional fine-tuning or annotation, with only the entity type definitions in the prompt adjusted accordingly. The results are presented in Table~\ref{Unseen-Datasets}.

\begin{table}[t]
\centering
\caption{Performance on Unseen Datasets}
\resizebox{\linewidth}{!}{
\begin{tabular}{lccccccc}
\hline
            & \multicolumn{3}{c}{CCKS2019-AS} &  & \multicolumn{3}{c}{NCBI-Disease} \\ \cline{2-4} \cline{6-8} 
            & P         & R        & F1       &  & P         & R         & F1       \\ \hline
GLM4-9B     & 38.48     & 26.83    & 31.61    &  & 28.42     & 63.12     & 39.19    \\
DeepSeek-R1 & \textbf{43.89}     & 39.21    & 41.42    &  & 63.14     & 62.87     & 63.01    \\
GPT-4       & 37.21     & 34.27    & 35.68    &  & 52.13     & 53.26     & 52.69    \\
Ours        & 38.36     & \textbf{46.96}    & \textbf{42.23}    &  & \textbf{71.83}     & \textbf{66.67}     & \textbf{69.15}    \\ \hline
\end{tabular}
}
\label{Unseen-Datasets}
\end{table}

The experimental results in Table~\ref{Unseen-Datasets} demonstrate that, compared to other general LLMs, the proposed multi-dataset fine-tuned model achieves leading F1 score on both unseen datasets, notably achieving 69.15\% on NCBI-Disease and significantly outperforming GPT-4 and DeepSeek-R1. This indicates a strong cross-task transfer ability. Furthermore, accurately conveying the target task’s entity definitions in the prompt proves essential for enabling zero-shot BioNER with LLMs, as it effectively guides the model to focus on the specific structural and semantic requirements of the new task, thereby unlocking its generalization potential.

\section{DISCUSSION}

To further validate the effectiveness of the proposed method, we conducted a manual analysis of the recognition differences among various models on the GENIA dataset, as shown in Table~\ref{Case-Studies}. The table compares predictions from the gold standard (GS), a model fine-tuned solely on the GENIA dataset (S-SFT), the best-performing existing baseline method on this dataset (DiFiNet), and our proposed framework (Ours).

\begin{table}[]
\centering
\caption{Case Studies of BioNER Predictions by Different Methods}
\resizebox{\linewidth}{!}{
\begin{tabular}{p{0.5cm}lll}
\hline
\multirow{5}{*}{Ex.1} &
\multicolumn{3}{l}{\makecell[l]{Input Text: “We have tested several subregions of the DR enhancer\\ B domain.”}} \\ \cline{2-4}
& GS     & DR enhancer (DNA) & B domain (DNA) \\
& S-SFT  & DR enhancer (DNA) & None \\
& DiFiNet & DR enhancer B domain (RNA) & B domain (DNA) \\
& Ours   & DR enhancer (DNA) & B domain (DNA) \\ \hline

\multirow{5}{*}{Ex.2} &
\multicolumn{3}{l}{\makecell[l]{Input Text: “...measured the concentration of VDR in PHA-activated\\ peripheral mononuclear cells.”}} \\ \cline{2-4}
& GS     & \multicolumn{2}{l}{PHA-activated peripheral mononuclear cells (Cell\_line)} \\
& S-SFT  & \multicolumn{2}{l}{None} \\
& DiFiNet & \multicolumn{2}{l}{None} \\
& Ours   & \multicolumn{2}{l}{None} \\ \hline
\end{tabular}
}
    \begin{tablenotes}
    	\item \textbf{Notes}: Entity types are shown in parentheses.
    \end{tablenotes}
\label{Case-Studies}
\end{table}

The analysis reveals notable differences between the original methods and our approach. In \textbf{Example 1} (Ex.1), the S-SFT model failed to identify the sub-entity “\textit{B-domain}”, while our method successfully recognized it, which is likely due to the complementary features learned through multi-dataset joint fine-tuning. Additionally, DiFiNet incorrectly identified the entire phrase “\textit{DR enhancer B domain}” as a single entity, whereas our framework accurately separated and retained two independent entities, “\textit{DR-enhancer}” and “\textit{B-domain}”, thereby demonstrating higher recognition precision.

Despite achieving strong overall results on the BioNER task, our framework still exhibits limitations. Similar to most existing methods, our approach exhibits weaknesses in semantic integration across sentence fragments or structural segments. This challenge makes it difficult to accurately identify complex biomedical entities involving long-range dependencies, as illustrated in \textbf{Example 2} (Ex.2). Specifically, our method fails to recognize the composite noun phrase “\textit{PHA-activated peripheral mononuclear cells}”, which reflects a common difficulty in BioNER when handling structurally complex expressions. To address these limitations, we plan to incorporate medical knowledge graphs to enhance semantic understanding. Aligning model outputs with predefined entities and relations during inference can help the model better capture complex biomedical entity descriptions.



\section{CONCLUSION}

In this study, we propose a unified BioNER framework tailored for LLMs. We reformulate the BioNER task as a text generation problem and design a symbolic tagging strategy to enable unified recognition of both flat and nested entities, with explicit boundary annotation. To improve the model’s generalization across languages and tasks, we adopt a bilingual Chinese-English multi-dataset joint fine-tuning strategy. Building on this foundation, we introduce a contrastive entity selector based on positive and negative sample construction to filter spurious or boundary-ambiguous predictions, thereby enhancing overall recognition precision. Experimental results on four benchmark datasets demonstrate that our approach achieves state-of-the-art performance in both Chinese and English BioNER tasks, surpassing traditional methods and existing LLM-based approaches. In future work, we plan to incorporate domain-specific knowledge graphs and integrate BioNER datasets in more languages to further enhance the model's semantic understanding and support broader biomedical applications in real-world clinical environments.

\section*{Acknowledgment}
This research was supported by the Natural Science Foundation of China (No. 62302076, 62276043), the Fundamental Research Funds for the Central Universities (No. DUT25YG108), and the Research Project on High Quality Development of Hospital Pharmacy, National Institute of Hospital Administration, NHC, China (No. NIHAYSZX2525).

\bibliographystyle{IEEEtran}

\begin{thebibliography}{10}
\providecommand{\url}[1]{#1}
\csname url@samestyle\endcsname
\providecommand{\newblock}{\relax}
\providecommand{\bibinfo}[2]{#2}
\providecommand{\BIBentrySTDinterwordspacing}{\spaceskip=0pt\relax}
\providecommand{\BIBentryALTinterwordstretchfactor}{4}
\providecommand{\BIBentryALTinterwordspacing}{\spaceskip=\fontdimen2\font plus
\BIBentryALTinterwordstretchfactor\fontdimen3\font minus \fontdimen4\font\relax}
\providecommand{\BIBforeignlanguage}[2]{{%
\expandafter\ifx\csname l@#1\endcsname\relax
\typeout{** WARNING: IEEEtran.bst: No hyphenation pattern has been}%
\typeout{** loaded for the language `#1'. Using the pattern for}%
\typeout{** the default language instead.}%
\else
\language=\csname l@#1\endcsname
\fi
#2}}
\providecommand{\BIBdecl}{\relax}
\BIBdecl

\bibitem{cariello2021comparison}
M.~C. Cariello, A.~Lenci, and R.~Mitkov, ``A comparison between named entity recognition models in the biomedical domain,'' in \emph{Proceedings of the Translation and Interpreting Technology Online Conference}, 2021, pp. 76--84.

\bibitem{tsuruoka2003boosting}
Y.~Tsuruoka and J.~Tsujii, ``Boosting precision and recall of dictionary-based protein name recognition,'' in \emph{Proceedings of the ACL 2003 workshop on Natural language processing in biomedicine}, 2003, pp. 41--48.

\bibitem{zhao2004named}
S.~Zhao, ``Named entity recognition in biomedical texts using an hmm model,'' in \emph{Proceedings of the international joint workshop on natural language processing in biomedicine and its applications (NLPBA/BioNLP)}, 2004, pp. 87--90.

\bibitem{makino2002tuning}
T.~Makino, Y.~Ohta, J.~Tsujii \emph{et~al.}, ``Tuning support vector machines for biomedical named entity recognition,'' in \emph{Proceedings of the ACL-02 workshop on Natural language processing in the biomedical domain}, 2002, pp. 1--8.

\bibitem{settles2005abner}
B.~Settles, ``Abner: an open source tool for automatically tagging genes, proteins and other entity names in text,'' \emph{Bioinformatics}, vol.~21, no.~14, pp. 3191--3192, 2005.

\bibitem{luo2020chinese}
L.~Luo, Z.~Yang, Y.~Song, N.~Li, and H.~Lin, ``Chinese clinical named entity recognition based on stroke elmo and multi-task learning,'' \emph{Chinese Journal of Computers}, vol.~43, no.~10, pp. 1943--1957, 2020.

\bibitem{luo2018attention}
L.~Luo, Z.~Yang, P.~Yang, Y.~Zhang, L.~Wang, H.~Lin, and J.~Wang, ``An attention-based bilstm-crf approach to document-level chemical named entity recognition,'' \emph{Bioinformatics}, vol.~34, no.~8, pp. 1381--1388, 2018.

\bibitem{zhang2018chinese}
Y.~Zhang and J.~Yang, ``Chinese ner using lattice lstm,'' \emph{arXiv preprint arXiv:1805.02023}, 2018.

\bibitem{lee2020biobert}
J.~Lee, W.~Yoon, S.~Kim, D.~Kim, S.~Kim, C.~H. So, and J.~Kang, ``Biobert: a pre-trained biomedical language representation model for biomedical text mining,'' \emph{Bioinformatics}, vol.~36, no.~4, pp. 1234--1240, 2020.

\bibitem{chen2022named}
P.~Chen, M.~Zhang, X.~Yu, and S.~Li, ``Named entity recognition of chinese electronic medical records based on a hybrid neural network and medical mc-bert,'' \emph{BMC medical informatics and decision making}, vol.~22, no.~1, p. 315, 2022.

\bibitem{li2022unified}
J.~Li, H.~Fei, J.~Liu, S.~Wu, M.~Zhang, C.~Teng, D.~Ji, and F.~Li, ``Unified named entity recognition as word-word relation classification,'' in \emph{proceedings of the AAAI conference on artificial intelligence}, vol.~36, no.~10, 2022, pp. 10\,965--10\,973.

\bibitem{yan2023embarrassingly}
H.~Yan, Y.~Sun, X.~Li, and X.~Qiu, ``An embarrassingly easy but strong baseline for nested named entity recognition,'' in \emph{Proceedings of the 61st Annual Meeting of the Association for Computational Linguistics (Volume 2: Short Papers)}, 2023, pp. 1442--1452.

\bibitem{cai2024difinet}
Y.~Cai, Q.~Liu, Y.~Gan, R.~Lin, C.~Li, X.~Liu, D.~Luo, and J.~JiayeYang, ``Difinet: Boundary-aware semantic differentiation and filtration network for nested named entity recognition,'' in \emph{Proceedings of the 62nd Annual Meeting of the Association for Computational Linguistics (Volume 1: Long Papers)}, 2024, pp. 6455--6471.

\bibitem{zhao2023chinese}
X.~Zhao, Z.~Shi, Y.~Xiang, and Y.~Ren, ``Chinese named entity recognition based on grid tagging and semantic segmentation,'' in \emph{2023 IEEE 9th International Conference on Cloud Computing and Intelligent Systems (CCIS)}.\hskip 1em plus 0.5em minus 0.4em\relax IEEE, 2023, pp. 289--294.

\bibitem{luo2023aioner}
L.~Luo, C.-H. Wei, P.-T. Lai, R.~Leaman, Q.~Chen, and Z.~Lu, ``Aioner: all-in-one scheme-based biomedical named entity recognition using deep learning,'' \emph{Bioinformatics}, vol.~39, no.~5, p. btad310, 2023.

\bibitem{luo2024taiyi}
L.~Luo, J.~Ning, Y.~Zhao, Z.~Wang, Z.~Ding, P.~Chen, W.~Fu, Q.~Han, G.~Xu, Y.~Qiu \emph{et~al.}, ``Taiyi: a bilingual fine-tuned large language model for diverse biomedical tasks,'' \emph{Journal of the American Medical Informatics Association}, vol.~31, no.~9, pp. 1865--1874, 2024.

\bibitem{zhou2023universalner}
W.~Zhou, S.~Zhang, Y.~Gu, M.~Chen, and H.~Poon, ``Universalner: Targeted distillation from large language models for open named entity recognition,'' \emph{arXiv preprint arXiv:2308.03279}, 2023.

\bibitem{wang2023instructuie}
X.~Wang, W.~Zhou, C.~Zu, H.~Xia, T.~Chen, Y.~Zhang, R.~Zheng, J.~Ye, Q.~Zhang, T.~Gui \emph{et~al.}, ``Instructuie: Multi-task instruction tuning for unified information extraction,'' \emph{arXiv preprint arXiv:2304.08085}, 2023.

\bibitem{dubey2024llama}
A.~Dubey, A.~Jauhri, A.~Pandey, A.~Kadian, A.~Al-Dahle, A.~Letman, A.~Mathur, A.~Schelten, A.~Yang, A.~Fan \emph{et~al.}, ``The llama 3 herd of models,'' \emph{arXiv e-prints}, pp. arXiv--2407, 2024.

\bibitem{glm2024chatglm}
T.~GLM, A.~Zeng, B.~Xu, B.~Wang, C.~Zhang, D.~Yin, D.~Zhang, D.~Rojas, G.~Feng, H.~Zhao \emph{et~al.}, ``Chatglm: A family of large language models from glm-130b to glm-4 all tools,'' \emph{arXiv preprint arXiv:2406.12793}, 2024.

\bibitem{yang2024qwen2}
A.~Yang, B.~Yang, B.~Zhang, B.~Hui, B.~Zheng, B.~Yu, C.~Li, D.~Liu, F.~Huang, H.~Wei \emph{et~al.}, ``Qwen2. 5 technical report,'' \emph{arXiv preprint arXiv:2412.15115}, 2024.

\bibitem{guo2025deepseek}
D.~Guo, D.~Yang, H.~Zhang, J.~Song, R.~Zhang, R.~Xu, Q.~Zhu, S.~Ma, P.~Wang, X.~Bi \emph{et~al.}, ``Deepseek-r1: Incentivizing reasoning capability in llms via reinforcement learning,'' \emph{arXiv preprint arXiv:2501.12948}, 2025.

\bibitem{keloth2024advancing}
V.~K. Keloth, Y.~Hu, Q.~Xie, X.~Peng, Y.~Wang, A.~Zheng, M.~Selek, K.~Raja, C.~H. Wei, Q.~Jin \emph{et~al.}, ``Advancing entity recognition in biomedicine via instruction tuning of large language models,'' \emph{Bioinformatics}, vol.~40, no.~4, p. btae163, 2024.

\bibitem{zhang2021cblue}
N.~Zhang, M.~Chen, Z.~Bi, X.~Liang, L.~Li, X.~Shang, K.~Yin, C.~Tan, J.~Xu, F.~Huang \emph{et~al.}, ``Cblue: A chinese biomedical language understanding evaluation benchmark,'' \emph{arXiv preprint arXiv:2106.08087}, 2021.

\bibitem{kim2003genia}
J.-D. Kim, T.~Ohta, Y.~Tateisi, and J.~Tsujii, ``Genia corpus—a semantically annotated corpus for bio-textmining,'' \emph{Bioinformatics}, vol.~19, no. suppl\_1, pp. i180--i182, 2003.

\bibitem{luo2022biored}
L.~Luo, P.-T. Lai, C.-H. Wei, C.~N. Arighi, and Z.~Lu, ``Biored: a rich biomedical relation extraction dataset,'' \emph{Briefings in Bioinformatics}, vol.~23, no.~5, p. bbac282, 2022.

\bibitem{li2016biocreative}
J.~Li, Y.~Sun, R.~J. Johnson, D.~Sciaky, C.-H. Wei, R.~Leaman, A.~P. Davis, C.~J. Mattingly, T.~C. Wiegers, and Z.~Lu, ``Biocreative v cdr task corpus: a resource for chemical disease relation extraction,'' \emph{Database}, vol. 2016, 2016.

\bibitem{dettmers2023qlora}
T.~Dettmers, A.~Pagnoni, A.~Holtzman, and L.~Zettlemoyer, ``Qlora: Efficient finetuning of quantized llms,'' \emph{Advances in neural information processing systems}, vol.~36, pp. 10\,088--10\,115, 2023.

\bibitem{han2020overview}
X.~Han, Z.~Wang, J.~Zhang, Q.~Wen, W.~Li, B.~Tang, Q.~Wang, Z.~Feng, Y.~Zhang, Y.~Lu \emph{et~al.}, ``Overview of the ccks 2019 knowledge graph evaluation track: entity, relation, event and qa,'' \emph{arXiv preprint arXiv:2003.03875}, 2020.

\bibitem{dougan2014ncbi}
R.~I. Do{\u{g}}an, R.~Leaman, and Z.~Lu, ``Ncbi disease corpus: a resource for disease name recognition and concept normalization,'' \emph{Journal of biomedical informatics}, vol.~47, pp. 1--10, 2014.

\end{thebibliography}

\vspace{12pt}


\end{document}